\documentclass[a4paper,fleqn,longmktitle]{cas-sc}
\usepackage{wrapfig}
\usepackage[numbers]{natbib}
\usepackage{setspace}
\usepackage{tikz}
\usepackage{graphicx}      
\usepackage{amsmath}
\usepackage{algorithm}
\usepackage{algorithmicx}
\usepackage{algpseudocode}
\usepackage[flushleft]{threeparttable}
\usepackage{lipsum}
\usepackage{multicol}
\usepackage{amssymb}
\usepackage{booktabs}
\usepackage{tikz}
\usepackage{hyperref}
\usepackage{ragged2e}
\usepackage{url}
\usetikzlibrary{positioning, fit, arrows.meta, shapes}
\usepackage{textcomp}
\usepackage{subfigure}
 \pdfoutput=1

\def\tsc#1{\csdef{#1}{\textsc{\lowercase{#1}}\xspace}}
\tsc{WGM}
\tsc{QE}
\tsc{EP}
\tsc{PMS}
\tsc{BEC}
\tsc{DE}

\begin{document}
\let\WriteBookmarks\relax
\def\floatpagepagefraction{1}
\def\textpagefraction{.001}
\shorttitle{}
\shortauthors{Agarwal et~al.}

\title [mode = title]{Hierarchical Deep Recurrent Neural Network based Method for Fault Detection and Diagnosis}                      
\tnotemark[1]


\author[1]{Piyush Agarwal}


\address[1]{Chemical Engineering Department, University of Waterloo, Ontario, Canada, N2L3G1}

\author[2]{Jorge Ivan Mireles Gonzalez}

\author[3]{Ali Elkamel}


\author[4]{Hector Budman}
\cormark[1]



\doublespacing
\begin{abstract}
A Deep Neural Network (DNN) based algorithm is proposed for the detection and classification of faults in industrial plants. The proposed algorithm has the ability to classify faults, especially incipient faults that are difficult to detect and diagnose with traditional threshold based statistical methods or by conventional Artificial Neural Networks (ANNs). The algorithm is based on a Supervised Deep Recurrent Autoencoder Neural Network (Supervised DRAE-NN) that uses dynamic information of the process along the time horizon. Based on this network a hierarchical structure is formulated by grouping faults based on their similarity into subsets of faults for detection and diagnosis. Further, an external pseudo-random binary signal (PRBS) is designed and injected into the system to identify incipient faults. The hierarchical structure based strategy improves the detection and classification accuracy significantly for both incipient and non-incipient faults. The proposed approach is tested on the benchmark Tennessee Eastman Process resulting in significant improvements in classification as compared to both multivariate linear model-based strategies and non-hierarchical nonlinear model-based strategies.
\end{abstract} 




\maketitle

\section{Introduction}

Process faults significantly impact the profit of chemical plants. A fault in a dynamic system is an anomalous variation that results in the deviation of process state variables from its acceptable range of operation \cite{isermann2005fault}. Since the effect of faults often propagates along the process it is imperative to detect them soon upon their occurrence. To mitigate the economic losses resulting from faults, industrial plants are often operated with multiple sensors and control loops that employ these sensors for feedback corrective action. However, in the presence of large process disturbances and manipulated variable constraints, these control schemes are not sufficiently resilient to avoid abnormal operation \cite{chiang2000fault}.\\

There are two different major approaches to fault detection and diagnosis (FDD) for industrial process systems namely, active and passive. Most of the work in the area of process systems engineering for FDD is based on passive approaches where the system outputs are monitored for detecting observable statistical changes. The active approach for FDD involves injecting persistently exciting input signal of specific bandwidth into the system and using the resulting input-output data for incipient fault detection and diagnosis \cite{heirung2019input,cusido2011signal,busch2014active}. In this work, a blend of both passive and active approaches are used where the passive approach is shown to be effective for identifying most faults but an active approach is required for detecting incipient faults. A fault is generally referred in the literature to as observable when its occurrence can be observed from a set of measured variables \cite{shams2011finding}. Observability/Diagnosibility is an important aspect in any fault detection and diagnosis problem since the lack of it leads to incorrect detection and miss-classification. Lack of observability often arises due to low signal to noise ratio in the measurements used for FDD and the presence of feedback control \cite{isermann2005fault}. The purpose of feedback controllers is partly to compensate for anomalous system variations which can mask the effects of certain faults. In addition, lack of distinguishability between different faults is related to the fact that different faults have a similar effect on the measured variables.\\

A typical process monitoring system is composed of two parts: a detection algorithm and a classification algorithm. The objective of detection is to make a binary decision on whether the process is in normal or faulty operation. After detecting abnormal operation, a fault classification algorithm is used to infer the type of fault and to determine which associated process variables are affected by the fault. In the current study, we simultaneously perform detection and classification with a single algorithm by considering the normal operation condition as an additional fault class to be identified in the classification step.\\

Process monitoring schemes rely on process models that are trained using historical data and then are used to infer faults. Based on the type of model used, the schemes can be classified into two main approaches: mechanistic  model-based (e.g. using first principles models) and data-driven model-based approaches \cite{chiang2000fault}. Data-driven models, such as the one used in the current study, are based on a comparison between combination of different sensor measurements corresponding to normal behaviour to the values of these variables corresponding to the faulty operation \cite{yin2014review}.\\

\begin{figure}
\begin{center}
       \includegraphics[width = 0.75\textwidth, height = 24em]{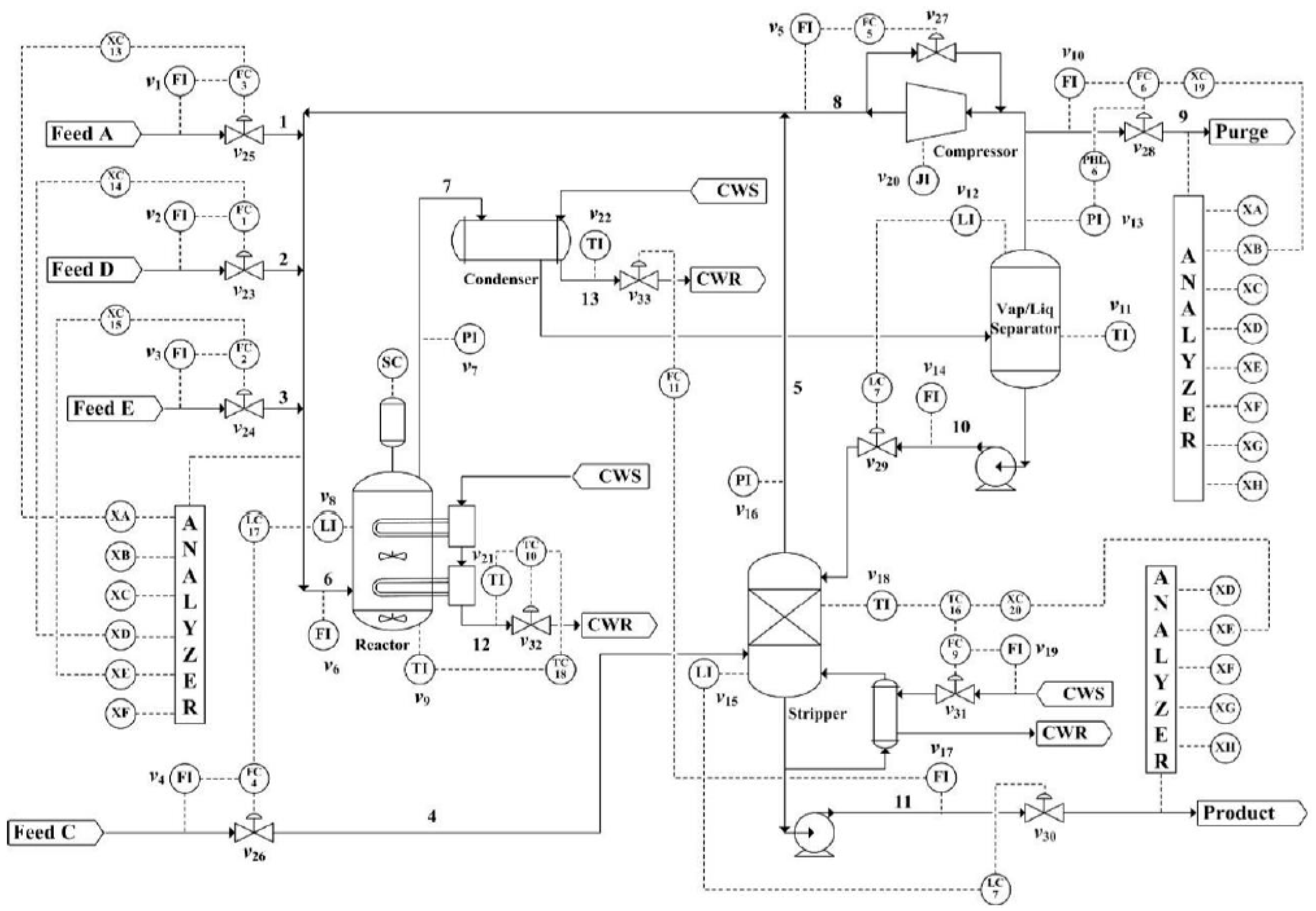}
	\caption{Schematic: Tennessee Eastman plant process\\(Downs \& Vogel, 1993)}
	\label{TEP_schematic} 
    \end{center}
\end{figure}

Within the class of data-driven approaches several algorithms predominantly are based on multivariate statistical methods such as PCA (Principal Component Analysis) \cite{zhang2009enhanced,yin2012comparison,lau2013fault,shams2010fault} or their dynamic variants such as DPCA (Dynamic Principal Component Analysis) \cite{chiang2000fault,yin2012comparison,ku1995disturbance,rato2013fault,odiowei2009nonlinear} have been proposed. Since the above mentioned methods were based on assumptions of process linearity, nonlinear modelling techniques such as ANN (Artificial Neural Network) based methods are investigated to deal with nonlinear process behavior. Some of the key challenges with the earlier versions of ANN algorithms were related to the difficulty in training large networks and perform complex calculations given the computational limitations that existed at the time when these algorithms were first proposed. In the last decade, a new generation of Deep Neural Networks (DNNs) algorithms has emerged that capitalizes both on the significant increase in computational power as well as novel algorithmic developments that facilitate the training and calibration of these networks. The use of these algorithms for fault detection in the process industry has recently received increased attention. However, despite the improvements in detection accuracy obtained with these techniques for nonlinear problems the classification of incipient faults remains a challenge. Based on the above facts, the current study focuses on deep learning techniques for the detection of faults with emphasis on incipient faults. Towards this goal, a hierarchical classification strategy based on DNN is proposed that involves identifying separate models for different subsets of faults with different signal to noise ratio characteristics. An addition of a test signal is also investigated to enhance fault diagnosibility for faults that are particularly difficult to identify. The proposed approach is based on dynamic network models that explicitly exploit the dynamic correlations in the data , i.e. auto-correlations and cross-correlations. In addition, the effects of data horizon or time length is also addressed.\\

All studies are conducted with the standard simulated data from the Tennessee Eastman Process (TEP). Since its introduction in the process systems research, the TEP has served as a benchmark problem for testing control and fault detection algorithms and it is thus ideal for comparing existing approaches to our proposed algorithm.
The following are the main contributions of the current study:\\
\begin{enumerate}
    \item Analysis of the effect of data horizon in the dynamic deep learning model to improve fault classification ability.
    \item Development of a hierarchical structure combined with the design of external excitation signals to enhance the detection and classification accuracy for both incipient and non-incipient faults.
    \item Comparison of the proposed algorithm to both linear multivariate statistical techniques and other deep learning (DL) based methods previously reported.
\end{enumerate}

The manuscript is organized as follows. Different fault detection and classification algorithms that are relevant to the TEP problem are briefly reviewed in Section 2. Section 3 presents the proposed methodology. Section 4 describes the  case study. The results and comparisons with previously reported approaches are presented in Section 5 followed by conclusions.

\section{Review of FDD Techniques relevant to the Tennessee Eastman Process (TEP)}

The Tennessee Eastman plant has been used widely for testing several process monitoring and fault detection algorithms \cite{chiang2000fault,lau2013fault,rato2013fault,xie2015hierarchical,ricker1996decentralized,bathelt2015revision,kulkarni2005knowledge,larsson2001self}. Thus, the current brief review of detection and diagnosis methods mainly focuses on TEP, that is also used in the current work as the case study. Also, some recently reported applications of deep learning algorithms that are relevant to the current study are included.\\

Figure \ref{TEP_schematic} shows the flow-sheet of the TEP process consisting of different interconnected unit operations including a vapor-liquid separator, a reactor, stripper a recycle compressor and a condenser. The simulation contains 20 pre-programmed fault scenarios, which are shown in Table \ref{process faults}. Additional details about the process model can be found in the original paper \cite{downs1993plant} and descriptions of the different control schemes that have been applied to the simulator can be found in \cite{ricker1996decentralized} and its revised version \cite{bathelt2015revision}. Several data-driven statistical process monitoring approaches have been reported for the detection and diagnosis of disturbances in the Tennessee Eastman simulation. Each of these methods has shown different levels of success in detecting and diagnosing the 20 faults considered in the simulations (Table 1).  Several statistical studies have reported faults 3, 9 and 15 as unobservable or difficult to diagnose due to the close similarity in the responses of the noisy measurements used to detect these faults \cite{lau2013fault,shams2010fault,chiang2000fault,du2018fault}. These 3 difficult to observe faults are referred hereafter as incipient faults.  It should be emphasized that the responses associated with incipient faults are similar but not identical to each other. Ideally in a noise free case, if a perfect model is available then these three faults could be correctly diagnosed. However, that is not the case in the presence of the noise levels used in the TEP studies and thus these incipient faults are incorrectly diagnosed most of the time.\\

Among the techniques used for detection, Principal Component Analysis (PCA) and its variants are widely employed \cite{zhang2009enhanced,yin2012comparison,lau2013fault}. PCA is an unsupervised learning data-driven technique based on orthogonal transformations that compresses a multivariate dataset into a lower-dimensional space while conserving the most relevant information \cite{pearson1901liii,hotelling1933analysis}. Extension of the PCA algorithm to enhance FDD involves the use of dynamic information since PCA capture static correlations only. For this purpose, Dynamic Principal Component Analysis (DPCA) was proposed that uses a dynamic data matrix to learn dynamic correlations \cite{ku1995disturbance}. Although DPCA improved the diagnosis accuracy over the results obtained with PCA on many TEP faults, it does not significantly improved the detection of the incipient faults \cite{chiang2000fault,ku1995disturbance,rato2013fault,odiowei2009nonlinear}. Another variant of the PCA approach that combines the results from the PCA algorithm with a Cumulative Sum (CUSUM) operation has shown to be a viable option to solve the detection problem of the incipient faults but a relative long time after occurrence of the fault is needed for detection. The reason is that the cumulative sum of PCA score values over a sufficient amount of time can provide detection of minor changes in the process variables that cannot be detected without using the CUSUM operation \cite{shams2010fault,shams2011finding}. \\

Typically, using the multivariate statistical algorithms reviewed above, it is possible to differentiate between normal operation to faulty operation, i.e. nominal operation versus faulty by comparing the values of the normal state with values of the state in the presence of the fault. Then, by combining the results of these detection algorithms with supervised classification techniques it is also possible to identify the particular fault occurring in the process. For example, Support Vector Machines (SVM) is a supervised-learning technique that is based on transforming the input data into a high dimensional space such as the distance between two different classes \cite{vapnik1999overview} is maximized. SVM has been applied to the TEP for both detection and diagnosis of faults \cite{kulkarni2005knowledge,chiang2004fault,mahadevan2009fault}.\\

Deep learning fault detection and classification techniques have been widely researched for applications in several engineering fields \cite{agarwal2019classification,agarwal2019deep}. In chemical engineering, machine learning techniques have been applied for the detection and classification of faults in the Syschem plant, which contains 19 different faults \cite{hoskins1991fault} and for the TEP problem. Outside the process industries, several studies on deep learning approaches have been conducted for the prevention of mechanical failures. For example deep learning  models have been used for detecting and diagnosing faults present in rotating machinery \cite{janssens2016convolutional,jia2016deep}, motors \cite{sun2016sparse}, wind turbines \cite{zhao2018anomaly}, rolling element bearings \cite{gan2016construction,he2017deep} and gearboxes \cite{jing2017convolutional,chen2015gearbox}. Few deep learning studies have been recently conducted on TEP. Similar to linear multivariate statistical approaches, deep learning methods for incipient faults have not significantly improved the diagnosis accuracy as compared to previous studies \cite{xie2015hierarchical,lv2016fault,zhao2018sequential}. \\

\citeauthor{xie2015hierarchical}, \citeyear{xie2015hierarchical} proposed neural network based methodology as a solution for the diagnosis problem in the Tennessee Eastman simulation that combines the network model with a clustering approach. The classification results obtained by this method were satisfactory for most non-incipient faults faults but were not satisfactory for incipient faults. Both Wang et al., \citeyear{wang2018generative} \cite{wang2018generative} and Spyridon et al. , \citeyear{spyridon2018generative} \cite{spyridon2018generative} proposed the use of Generative Adversarial Networks (GANs), as a fault detection scheme for the TEP. GANs are an unsupervised technique composed of a generator and a discriminator trained with the adversarial learning mechanism, where the generator replicates the normal process behavior and the discriminator decides if there is abnormal behavior present in the data. This unsupervised technique can detect changes in the normal behavior achieving good detection rates for non incipient faults. Lv et al., \citeyear{lv2016fault} \cite{lv2016fault} proposed a stacked sparse autoencoder (SSAE) structure with a deep neural network to extract important features from the input to improve the diagnosis problem in the Tennessee Eastman simulation. The diagnosis results applying this deep learning technique showed improvements compared to other linear and non-linear methods for non-incipient faults. To account for dynamic correlations in the data, Long Short Term Memory (LSTM) units have been recently applied to the TEP for the diagnosis of faults \cite{zhao2018sequential}. A model with LSTM units was used to learn the dynamical behaviour from sequences and batch normalization was applied to enhance convergence. An alternative to capture dynamic correlations in the data is to apply a Deep Convolutional Neural Networks (DCNN) composed of convolutional layers and pooling layers \cite{wu2018deep}. A DCNN model was constructed to learn the dynamic behaviour of different faults by taking advantage of the spatial, i.e.  feature space, and temporal domains. While this proposed deep learning algorithm was shown to achieve high classification rates for non-incipient faults it was not accurate for incipient faults. \\

Following the above, the current study investigates a different model structure for detecting and diagnosing faults with particular focus on the  detection and classification of incipient faults by the following means: i- extension of the time horizon used in the LSTM network, ii- the use of a hierarchical structure that uses separate models for incipient and non-incipient faults respectively and iii- the design and injection of external PRBS excitation signal.

\begin{table}[width=\linewidth,cols=3,pos=H]
\centering
\caption{Process Faults for classification in TE Process}
\label{process faults}
\resizebox{0.6\textwidth}{!}{%
\begin{tabular}{@{}llc@{}}
\toprule
\multicolumn{1}{l}{Fault} & \multicolumn{1}{c}{Description} & Type \\ \midrule
IDV(1)	& A/C feed ratio, B composition constant (stream 4)	& step\\
IDV(2)	& B composition, A/C ratio constant (stream 4)	& step\\
IDV(3) & D Feed Temperature & step\\
IDV(4)	& Reactor cooling water inlet temperature	& step\\
IDV(5)	& Condenser cooling water inlet temperature (stream 2) & step\\
IDV(6) &	A feed loss (stream 1) & step\\
IDV(7) &	C header pressure loss reduced availability (stream 4)	& step\\
IDV(8) &	A, B, C feed composition (stream 4)	& random variation\\
IDV(9) & D Feed Temperature & random variation\\
IDV(10) & C feed temperature (stream 4)	& random variation\\
IDV(11) & Reactor cooling water inlet temperature & random variation\\
IDV(12)	& Condenser cooling water inlet temperature	 & random variation\\
IDV(13) & Reaction kinetics & 	slow drift\\
IDV(14)	& Reactor cooling water & valve	sticking\\
IDV(15) & Condenser Cooling Water Valve & stiction\\
IDV(16) & Deviations of heat transfer within stripper & random variation\\
IDV(17) & Deviations of heat transfer within reactor & random variation\\
IDV(18) & Deviations of heat transfer within condenser & random variation\\
IDV(19) & Recycle valve of compressor, underflow stripper and steam valve stripper & stiction\\
IDV(20) & unknown & random variation \\
\bottomrule
\end{tabular}}
\end{table}

\begin{table}[width=\linewidth,cols=3,pos=H]
\centering
\caption{Measured and manipulated variables (from Downs and Vogel, 1993)}
\label{measured_outputs}
\resizebox{\textwidth}{!}{%
\begin{tabular}{@{}lcclcc@{}}
\toprule
\multicolumn{1}{l}{Variable Name} & \multicolumn{1}{c}{Variable Number} & Units & \multicolumn{1}{l}{Variable Name} & \multicolumn{1}{c}{Variable Number} & Units\\ \midrule
A feed (stream 1) & XMEAS (1) & kscmh & Reactor cooling water outlet temperature & XMEAS (21) & $^\circ$ C\\
D feed (stream 2) & XMEAS (2) & kg h$^{-1}$ & Separator cooling water outlet temperature & XMEAS (22) & $^\circ$C \\
E feed (stream 3) & XMEAS (3) & kg h$^{-1}$ & Feed \%A &  XMEAS(23) &  mol\% \\ 
A and C feed (stream 4) & XMEAS (4) & kscmh & Feed \%B &  XMEAS(24) &  mol\% \\ 
Recycle flow (stream 8) & XMEAS (5) & kscmh & Feed \%C &  XMEAS(25) &  mol\% \\ 
Reactor feed rate (stream 6) & XMEAS (6) & kscmh & Feed \%D &  XMEAS(26) &  mol\% \\ 
Reactor pressure & XMEAS (7) & kPa guage & Feed \%E &  XMEAS(27) &  mol\% \\
Reactor level & XMEAS (8) & \% & Feed \%F&  XMEAS(28) &  mol\% \\
Reactor temperature & XMEAS (9) & $^\circ$C &  Purge \%A&  XMEAS(29) &  mol\%\\ 
Purge rate (stream 9) & XMEAS (10) & kscmh & Purge \%B &  XMEAS(30) &  mol\%\\
Product separator temperature & XMEAS (11) & $^\circ$C & Purge \%C &  XMEAS(31) &  mol\%\\
Product separator level & XMEAS (12) & \% & Purge \%D &  XMEAS(32) &  mol\%\\
Product separator pressure & XMEAS (13) & kPa guage & Purge \%E &  XMEAS(33) &  mol\%\\
Product separator underflow (stream 10) & XMEAS (14) & m$^3$ h$^{-1}$ & Purge \%F &  XMEAS(34) &  mol\%\\
Stripper level & XMEAS (15) & \% & Purge \%G &  XMEAS(35) &  mol\%\\
Stripper pressure & XMEAS (16) & kPa guage & Purge \%H &  XMEAS(36) &  mol\%\\
Stripper underflow (stream 11) & XMEAS (17) & m$^3$ h$^{-1}$ & Product \%D & XMEAS(37) & mol\%\\
Stripper temperature & XMEAS (18) & $^\circ$C & Product \%E & XMEAS(38) & mol\%\\
Stripper steam flow & XMEAS (19) & kg h$^{-1}$ & Product \%F & XMEAS(39) & mol\%\\
Compressor Work & XMEAS (20) & kW & Product \%G & XMEAS(40) & mol\%\\
D Feed Flow & XMV (1) & kg h$^{-1}$& Product \%H & XMEAS(41) & mol\% \\
E Feed Flow & XMV (2) & kg h$^{-1}$ & 
A Feed Flow & XMV (3) & kscmh\\
A + C Feed Flow & XMV (4) & kscmh & Compressor Recycle Valve & XMV(5) & \%\\
Purge Valve & XMV (6) & \% &
Separator pot liquid flow & XMV (7) & m$^{3}$h$^{-1}$\\
Stripper liquid product flow & XMV (8) & m$^{3}$h$^{-1}$ & Stripper Steam Valve & XMV (9) & \%\\
Reactor cooling water flow & XMV (10) & m$^{3}$h$^{-1}$ &
Condenser cooling water flow & XMV (11) & m$^{3}$h$^{-1}$\\
\bottomrule
\end{tabular}}
\end{table}

\section{Preliminaries}
\subsection{Recurrent Neural Networks (RNNs)}

The current study uses a Recurrent Neural Network (RNN) model that was originally developed for handling dynamic data by using time sequences of data $\textbf{x}_t^{~i},$ $t = 1,2,...,T \in \mathbb{R}^{d_h \times d_x}$. \cite{rumelhart1986learning} as inputs to the network. Parameters associated with RNN are shared along a time horizon to capture temporal correlations in data. This enhances the generalization capability of the model to time sequences that were not used for model calibration. Figure 2 shows a schematic description of a simple example of recurrent neural networks. As shown in Figure 2, recurrent models are composed of feed-forward connections, which represent the flow of information from a neuron to another($\textbf{w}_{in}$ and $\textbf{w}_{out}$), and recurrent connections, which captures important information stored from previous time steps($\textbf{w}_{rec}$). Figure 2, shows a fully connected recurrent model that produces an output at each time step and contains a recurrent connection in its hidden layer. In this case, the equation of the hidden layer is formulated to account for the recurrent relation as follows: 

\begin{equation}
    \textbf{h}(t) = f(\textbf{z}) = f(\textbf{w}_{in}\textbf{x}^{(t)} + \textbf{w}_{rec}\textbf{h}^{(t-1)} + \textbf{b}) 
\end{equation}
A well-known challenge for training RNNs is the vanishing gradient or exploding gradient problem arising from the use of gradient descent algorithms in combination with sigmoid activation functions  \cite{bengio1994learning}. To deal with this problem, the best practice is to use gated-type unit structures within RNN models such as Long-Short Term Memory units (LSTM) \cite{hochreiter1997long} and Gated Recurrent Units (GRU) \cite{cho2014learning}. LSTM is reviewed in the following section since they serve as the basis for the models used in the current study for FDD.

\subsection{Long-Short Term Memory (LSTM) Units}
The LSTM unit is composed of three gated units and a memory cell \cite{hochreiter1997long}. Figure \ref{LSTM_cell} shows a single LSTM unit that includes four major gates:  the forget gate ($\textbf{f}_t$), the input gate ($\textbf{i}_t$), the output gate ($\textbf{o}_t$) and the update gate ($\textbf{g}_t$). The key component of the LSTM unit is the memory cell ($\textbf{c}_t \in \mathbb{R}^{d_h \times 1}$) that is responsible for storing critical long term dependencies learned over time. The input gate ($\textbf{i}_t$) is responsible for evaluating which part, if any, of the past historical data should be kept. Thus the function of the input gate is to allow the network to keep only relevant information from the previous time steps and discard the rest for a sample $i$. \\

Subsequently, the information that is worth recording is determined by the memory cell ($\textbf{c}_t$). The process of identifying information and storing in the memory cell consists of two parts: new information that is recorded and information that is discarded. The information that should be discarded from previous cell state $\textbf{c}_{t-1}^{i}$ is determined by the forget gate ($\textbf{f}_t$), which is responsible for forgetting previously stored cell state values that have lost their relevance. Then new relevant information is added and existing cell-state values are updated by first selecting which values to update using the input gate $\textbf{i}_t^{~i}$ and the output from the input gate is then multiplied by the new information generated by the update gate $\textbf{g}_t^{~i}$. Ultimately, the output $\textbf{h}_t$ is computed at every time step from the information contained in the memory cell and it is further gated by an output gate according to its importance or relevance. The mathematical equations describing these gating operations are as follows:
{\tiny\begin{figure}
\centering
\begin{tikzpicture}[
font=\sf \scriptsize,
>=LaTeX,
cell/.style={
	rectangle, 
	rounded corners=5mm, 
	draw,
	very thick,
},
operator/.style={
	circle,
	draw,
	inner sep=-0.5pt,
	minimum height =.2cm,
},
function/.style={
	ellipse,
	draw,
	inner sep=1pt
},
ct/.style={
	circle,
	draw,
	line width = .75pt,
	minimum width=1cm,
	inner sep=1pt,
},
gt/.style={
	rectangle,
	draw,
	minimum width=4mm,
	minimum height=3mm,
	inner sep=1pt
},
mylabel/.style={
	font=\scriptsize\sffamily
},
ArrowC1/.style={
	rounded corners=.25cm,
	thick,
},
ArrowC2/.style={
	rounded corners=.5cm,
	thick,
},
]
\node[] at (-2.5,0) {Forget};
\node[] at (-2.5,-0.2) {Gate};
\node[] at (-2.5,-0.44) {$\textbf{f}_t$};

\node[] at (-1.3,0.6) {Update};
\node[] at (-1.3,0.37) {Gate};
\node[] at (-1.3,0.13) {$\textbf{g}_t$};

\node[] at (-.0,-0.05) {Input};
\node[] at (-.0,-0.25) {Gate};
\node[] at (-.0,-0.48) {$\textbf{i}_t$};

\node[] at (.7,0.6) {Output};
\node[] at (.7,0.37) {Gate};
\node[] at (.7,0.13) {$\textbf{o}_t$};

\node [cell, minimum height =4cm, minimum width=6cm] at (0,0){} ;

\node [gt] (ibox1) at (-2,-0.75) {$\sigma$};
\node [gt] (ibox2) at (-1.5,-0.75) {$\sigma$};
\node [gt, minimum width=1cm] (ibox3) at (-0.5,-0.75) {tanh};
\node [gt] (ibox4) at (0.5,-0.75) {$\sigma$};

\node [operator] (mux1) at (-2,1.5) {$\times$};
\node [operator] (add1) at (-0.5,1.5) {+};
\node [operator] (mux2) at (-0.5,0) {$\times$};
\node [operator] (mux3) at (1.5,0) {$\times$};
\node [function] (func1) at (1.5,0.75) {tanh};

\node[ct, label={[mylabel]Cell}] (c) at (-4,1.5) {${c}_{t-1}$};
\node[ct, label={[mylabel]Hidden}] (h) at (-4,-1.5) {${h}_{t-1}$};
\node[ct, label={[mylabel]left:Input}] (x) at (-2.5,-3) {${x}_{t}$};

\node[ct, label={[mylabel]}] (c2) at (4,1.5) {${c}_{t}$};
\node[ct, label={[mylabel]}] (h2) at (4,-1.5) {${h}_{t}$};
\node[ct, label={[mylabel]left:}] (x2) at (2.5,3) {${h}_{t}$};

\draw [ArrowC1] (c) -- (mux1) -- (add1) -- (c2);

\draw [ArrowC2] (h) -| (ibox4);
\draw [ArrowC1] (h -| ibox1)++(-0.5,0) -| (ibox1); 
\draw [ArrowC1] (h -| ibox2)++(-0.5,0) -| (ibox2);
\draw [ArrowC1] (h -| ibox3)++(-0.5,0) -| (ibox3);
\draw [ArrowC1] (x) -- (x |- h)-| (ibox3);

\draw [->, ArrowC2] (ibox1) -- (mux1);
\draw [->, ArrowC2] (ibox2) |- (mux2);
\draw [->, ArrowC2] (ibox3) -- (mux2);
\draw [->, ArrowC2] (ibox4) |- (mux3);
\draw [->, ArrowC2] (mux2) -- (add1);
\draw [->, ArrowC1] (add1 -| func1)++(-0.5,0) -| (func1);
\draw [->, ArrowC2] (func1) -- (mux3);

\draw [-, ArrowC2] (mux3) |- (h2);
\draw (c2 -| x2) ++(0,-0.1) coordinate (i1);
\draw [-, ArrowC2] (h2 -| x2)++(-0.5,0) -| (i1);
\draw [-, ArrowC2] (i1)++(0,0.2) -- (x2);
\end{tikzpicture}
\caption{Schematic of a LSTM memory cell}
\label{LSTM_cell}
\end{figure}
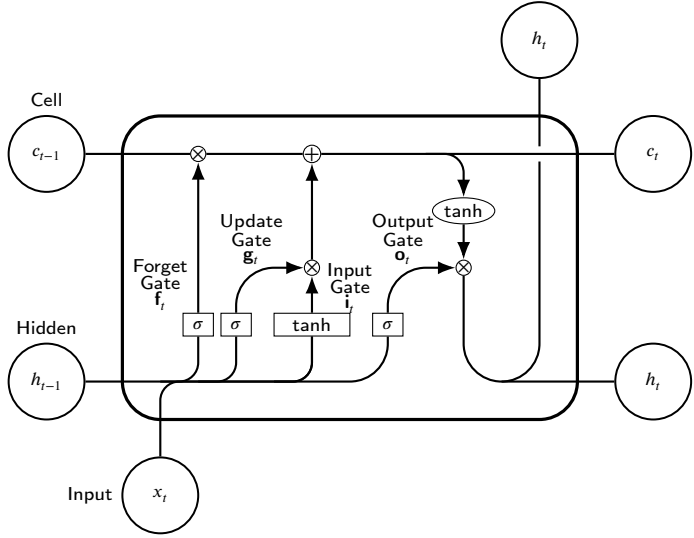}
    
\begin{align}
\nonumber \textbf{i}_t^{~i} &= \sigma(\textbf{W}_i\textbf{x}_t^{~i} + \textbf{R}_i\textbf{h}_{t-1}^{~i} + \textbf{b}_i)\\ 
	\textbf{g}_t^{~i} &= \text{tanh}(\textbf{W}_g\textbf{x}_t^{~i} + \textbf{R}_g\textbf{h}_{t-1}^{~i} + \textbf{b}_g)
	\label{itgt}
\end{align}
\begin{align}
    \textbf{c}_t^{~i} &= \textbf{f}_t^{~i}\odot\textbf{c}_{t-1}^{~i} + \textbf{i}_t^{~i} \odot \textbf{g}_t{~i}
    \label{ct}
\end{align}
where $\bf{\sigma}()$ and $\bf{\tanh}()$ are the element-wise sigmoid and hyperbolic tangent functions respectively.
\begin{align}
\nonumber	
	\textbf{o}_t^{~i} &= \sigma(\textbf{W}_o\textbf{x}_t^{~i} + \textbf{R}_o\textbf{h}_{t-1}^{~i} + \textbf{b}_o)\\
	\textbf{h}_t^{~i} &= \textbf{o}_t^{~i} \odot \text{tanh}(\textbf{c}_t^{~i})
	\label{otht}
\end{align}
where $\textbf{R}  =  [\textbf{R}_f~\textbf{R}_i~ \textbf{R}_g~\textbf{R}_o]^T \in \mathbb{R}^{4d_h\times d_h}$ are known as recurrent weights, $\textbf{W} = [\textbf{W}_f~\textbf{W}_i~ \textbf{W}_g~\\ \textbf{W}_o]^T \in \mathbb{R}^{4d_h\times d_x}$ are all the input weights, $\textbf{b} =[\textbf{b}_f~\textbf{b}_i~ \textbf{b}_g~\textbf{b}_o]^T \in \mathbb{R}^{d_h \times 1}$ are the bias parameters.

\subsection{Deep LSTM Supervised Autoencoder Neural Network (LSTM-SAE NN)}
\begin{figure}[]
    \centering
    \includegraphics[scale = 0.35 ]{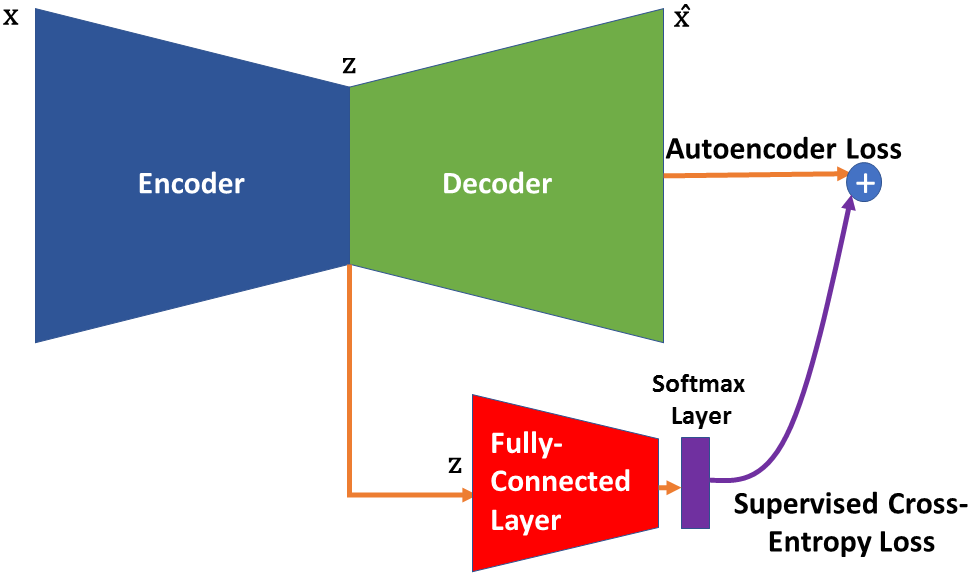}
    \caption{Schematic of a single layer Supervised Autoencoder Neural Network (SAE-NN)}
    \label{Schematic SAE}
\end{figure}

The training of an Deep Supervised Autoencoder Neural Network (DSAE-NN) model, schematically shown in Figure \ref{Schematic SAE}, is based on the minimization of a weighted sum of the reconstruction loss function and the supervised classification loss corresponding to the first and second terms in  (Equation (\ref{SAE_lossfunction})) respectively. The minimization of the reconstruction loss function in Equation (\ref{SAE_lossfunction}) ensures that the estimated latent variables are able to capture the variance in the input data while the minimization of the classification loss is ensures the non-linear latent variables extracted are the predictors of the output label. The mean squared error function is used as a reconstruction loss and softmax cross-entropy as the classification loss. The overall goal is to learn a function that predicts the class labels in one-hot encoded form $\textbf{y}_{i} \in \mathbb{R}^m$ from inputs $\textbf{x}_i \in \mathbb{R}^{d_x\times 1}$. In this work, we use LSTM units instead of dense layers for both the encoder and decoder. The goal is to reconstruct and classify input sequences at time $t$ simultaneously. The encoder transforms the input time sequences using the Equations \ref{itgt},\ref{ct} and \ref{otht} to learn important features and encode these features $\textbf{z} \in \mathbb{R}^{d_h \times 1}$. The decoder function reconstruct the input using the extracted feature vectors. The operation performed by the encoder for a single LSTM layer between the input variables to the latent variables $\mathbf{z}_t^i \in \mathbb{R}^{d_h \times 1}$ can be mathematically described as follows:

\begin{align}
    \textbf{z}_t^i = \zeta_e(\textbf{x}_t^i)
\end{align}

where $zeta_e$ is the LSTM encoder function The latent variables are used both to predict the class labels and to reconstruct back the inputs $\textbf{x}$ as follows:
\begin{align}
    \hat{\textbf{x}_t^i} = \zeta_d(z_t^i) \\
    \hat{\textbf{y}_t^i} = f_c(\textbf{W}_c\textbf{z}_t^i + \textbf{b}_c)
\end{align}

where $f_c$ is a non-linear activation function applied for the output layer. $\textbf{W}_c \in \mathbb{R}^{m \times d_z}$ and $\textbf{b}_c \in \mathbb{R}^{m}$ are output weight matrix and bias vector respectively. For training the SAE, the following loss function is minimized:
\begin{align} \nonumber l_{DSAE} &= 
  \frac{\lambda_1}{N}|| \textbf{x}_s - \hat{\textbf{x}}_s||_2^2 + \frac{1}{N}\sum_{s=1}^{N}\sum_{c=1}^{m} -y_{s,c}log(p_{s,c})\\ 
 &=  \frac{1}{N}\bigg[\lambda_1||\textbf{x}_s - \hat{\textbf{x}}_s||_2^2 +\sum_{s=1}^{N}\sum_{c=1}^{m} -y_{s,c}log(p_{s,c})\bigg]    
\label{SAE_lossfunction}
\end{align}

\begin{align}
    p_{s,c} &= \frac{e^{(\hat{y_{s,c}})}}{\sum_{c = 1}^{m}e^{(\hat{y_{s,c}})}}
\end{align}

where $\lambda_1$ is the weight multiplying the reconstruction loss $L_r$ in the cost to be minimized, $m$ is the number of classes, $y_{s,c}$ is a binary indicator (0 or 1) equal to 1 if the class label $c$ is the correct one for observation $s$ and 0 otherwise, $\hat{y_{s,c}}$ is the non-normalized log probabilities and $p_{s,c}$ is the predicted probability for a sample $s$ of class $c$. Moreover, to avoid over-fitting, a regularization term is added to the objective function in Equation \ref{SAE_lossfunction}. Accordingly, the objective function for Deep Supervised LSTM NNs used for FDD is as follows:
\begin{align}
    \min_{\textbf{W}} l_{DSAE} = \min ~~\frac{1}{N}\bigg[\lambda_1||\textbf{x}_s - \hat{\textbf{x}}_s||_2^2 +\lambda_2\sum_{s=1}^{N}\sum_{c=1}^{m} -y_{s,c}log(p_{s,c}) + \lambda_3\sum_{L}\sum_{k}\sum_{j}\textbf{W}^{[L]}_{{kj}^2} \bigg] 
    \label{DSAE_objective_function}
\end{align}

\begin{figure}[]
    \centering

    \includegraphics[scale = 0.35 ]{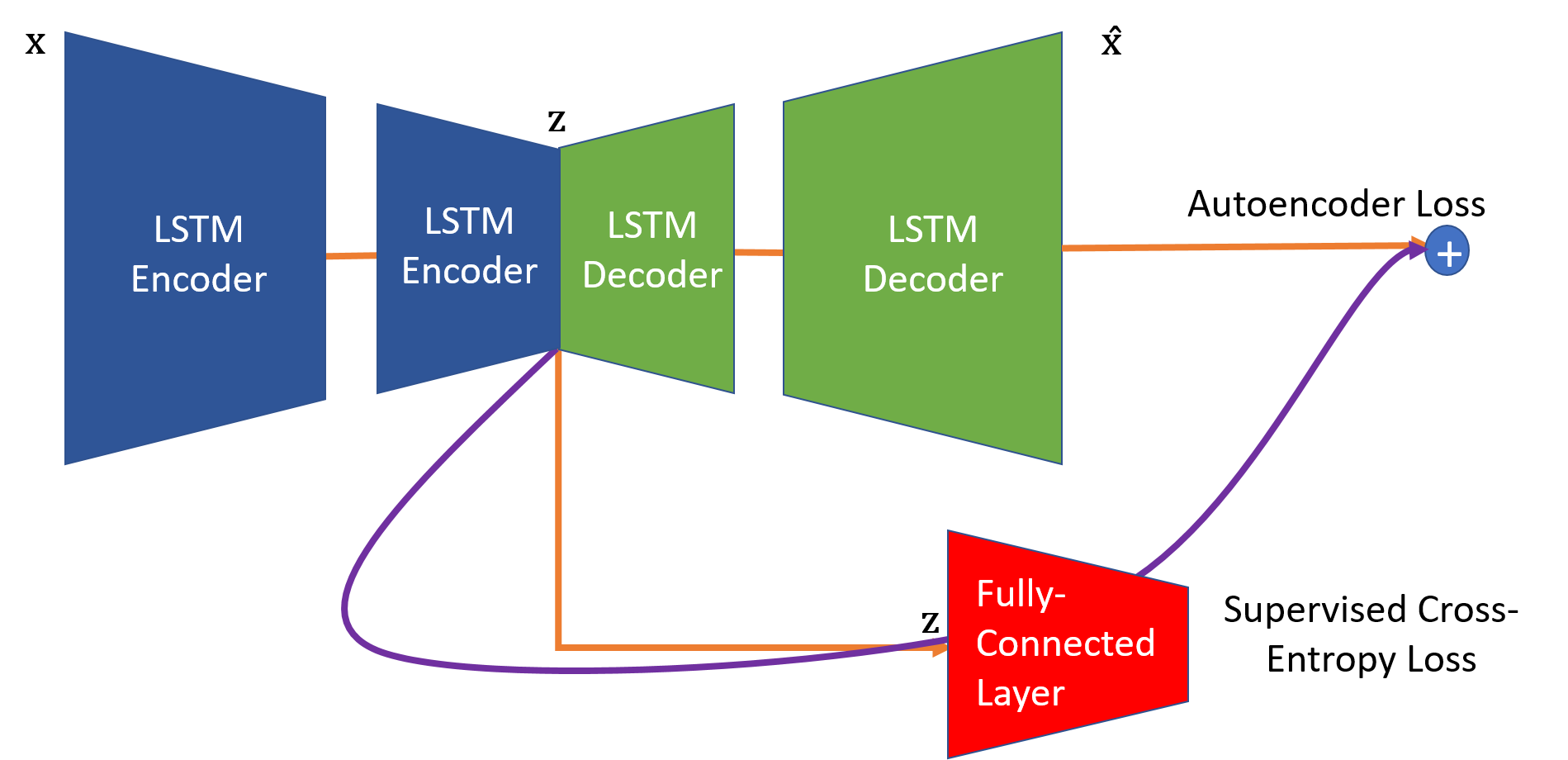}
    \caption{Schematic of a Deep LSTM Supervised Autoencoder Neural Network (DLSTM-SAE NN)}
    \label{Schematic LSTM-SAE}
\end{figure}

where $\textbf{W}_{kj}^{[L]}$ are the weight matrices for each layer $L$ in the network and the weights on the individual objective functions  $\lambda_1, \lambda_2, \lambda_3$ are chosen using validation data.

\subsection{Model Structure and Specifications}

The RNN based model with LSTM units used in the current study was developed with training and testing data sets generated from the Tennessee Eastman Process (TEP) simulation. The data are extracted from simulations of the system conducted at either the normal state or when each of the 20 different faults is occurring in the process. It is assumed that at each sampling interval, 52 different variables are measured and organized into a vector. Each such vector of measurements is acquired every 3 minutes. It should be noticed that during testing of the methods proposed in this study the normal state is considered as a different separate class and hence a total of 21 different classes, i.e. 20 faulty plus one normal operations, are considered for classification. The standard dataset can be downloaded from \url{http://depts.washington.edu/control/LARRY/TE/download.html}. The simulator is ran for 72 hours (training: 24 hours; testing: 48 hours) for each fault generating 1440 samples for each fault class and normal class. The data is then divided between calibration and validation data sets, where the first 480 samples are used as training data and the rest are used for testing for each class. This results in a total of 10,080 training samples and 19,200 testing samples. A small fraction of training dataset is used as validation dataset for selecting the optimal hyper-parameters. It is important to note that the number of training, validation and testing samples vary depending on the time horizon used in dynamical RNN based model. The results reported in the following section are based on the classification accuracy of test dataset, i.e. on data that was not used for model calibration. The experiments in this paper have been implemented on an Intel Core i7-7700HQ PC (2.80GHz, 16GB RAM) and NVIDIA GeForce GTX 1060 (6GB) 64Bit Windows 10 operating system in Python environment. The models are developed using Keras \cite{chollet2015keras} (an open deep learning library) on TensorFlow platform. \cite{abadi2016tensorflow}. All hyper-parameters such as number of LSTM encoder layers, LSTM units in each layer, weights and learning rate are optimized using Keras-tuner developed by Google team and is included in the Keras library.\\

\section{Hierarchical Structure}
\begin{figure}[]
    \centering
    \includegraphics[scale = 0.8]{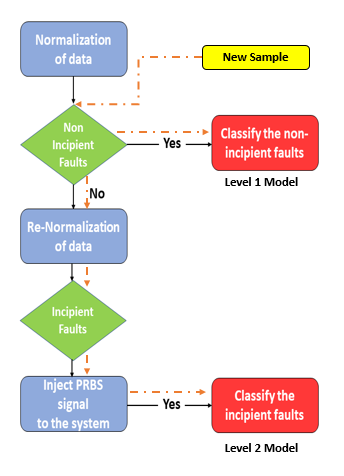}
    \caption{Hierarchical structure used for fault detection and diagnosis}
    \label{hierarchical}
\end{figure}

The sensitivity of nonlinear models such as deep neural networks is highly dependent on the variability of the data used for calibration. Accordingly, a key data pre-processing step towards model calibration involves data standardization. It is required to account for different ranges of values and engineering units of the inputs. It is hypothesized that by building separate models for different groups of faults based on similar characteristics of certain faults, it is possible to increase the sensitivity of different models and diagnosibility between faults because of the different specific data re-normalization step conducted within each group. Accordingly, a hierarchical structure is proposed in the current study as shown in Figure \ref{hierarchical}. In the first level of the hierarchical structure, the normal state condition is grouped along with incipient faults as class 1 and is classified against all the other non-incipient faults. Subsequently, in a second step, PRBS signals are injected into the system to distinguish between different incipient faults and normal state. This hierarchical structure can be summarized as follows: 
\begin{enumerate}
    \item A first-level hierarchical model is used to identify the non incipient faults 
    \item A second-level hierarchical model that focuses on the normal state and 3 difficult to observe faults, i.e. faults 3, 9 and 15. If the incipient faults cannot be properly identified in the second step then we inject the PRBS.
    \item If a new sample is not identified by the first level model then the data is analyzed by the second-level model and the fault is identified among the possible candidates, i.e. normal state, fault 3, 9 and 15.
\end{enumerate} 

It should be noted that the incipient faults are characterized by responses that are very similar to the normal state and thus a model that is trained to predict all the faults together will be shown to be unable to accurately discern between these responses. It should also be noted that incipient faults that are grouped along with normal state in the first level may also be miss-classified as other faults. Hence, the overall classification accuracy for the incipient faults after the second level is executed has to be re-calculated accordingly.\\

 In the hierarchical structure described in Figure 5 the normalized data is fed to a first level model where the softmax layer of LSTM-SAE NN uses 18 units instead of the 21 units (incipient faults and normal state grouped as one) as used in the non-hierarchical type model. The structure of the model in the second level of the hierarchical structure is similar to the first level one but the difference is that the softmax layer involves only 4 units each for one of the incipient faults (3,9,15) and for the normal state (fault 0). The PRBS is injected only when the incipient fault cannot be properly identified with the Hierarchical Deep RNN based model.

\subsection{Design: Pseudo-random Binary Signal (PRBS)}

Although the hierarchical structure proposed in the previous section enhances the diagnosibility of few faults, detection of incipient faults is still challenging due to lack of excitation to detect these faults in the presence of noise. This problem is particularly acute in the TEP since the data-set contains variables that are used in closed-loop control thus exhibiting small variation with respect to their set-point values making it difficult to estimate the occurrence of faults from such variables. To increase diagnosibility of incipient faults the use of active fault detection, as reviewed in the Introduction, is proposed for the TEP process. The lack of diagnosibility/distinguishability of the incipient faults can be viewed as a problem of inaccurate identification of a model relating variability in measured values to faults. To improve the identification accuracy it is required to use inputs that sufficiently excite the system dynamics in the presence of noise \cite{ljung1999system} which will result in larger changes in the measured quantities and larger sensitivity to fault changes. Thus, it is required to introduce additional excitation to the one available in regular operation of the system. Accordingly, external forcing signals are injected at particular points of the control loops, e.g. an excitation signal to the set-points of the loops that involve variables related to the difficult to detect faults. The addition of such excitation signals in combination with a separate deep neural network model (second level) in the hierarchical structure described in the previous section is investigated in the current study for detecting and diagnosing incipient faults that cannot be accurately identified with the regular operating data collected from the process.\\

To avoid a large negative impact of the external signals on the profitability of the plant the input signals should meet certain constraints as follows:

\begin{enumerate}
    \item Reduce input move sizes (to reduce wear and tear on actuators).
    \item  Reduce input and output amplitudes, power, or variance.
    \item Short experimental time to prevent losses
\end{enumerate}

In a practical implementation, the added excitation signal should result in variations in the measured quantities that  will be large in magnitude relative to the noise.  Towards this goal it is necessary to include information of frequencies lower than the crossover frequency of the closed loop transfer function \cite{rivera1995systematic}. PRBS signals are used as excitation signals in this study since they have a finite length that can be synthesized repeatedly with simple generators while presenting favorable spectra. The spectrum at low frequencies are flat and constant while at high frequencies the spectra drop off. Thus, the PRBS can be designed to have a specific bandwidth, which can be utilized for exciting the processes within the required range of
frequencies \cite{garciainput}. The analytical expression for the power spectrum of a PRBS is given by:
\begin{align}
s(\omega) = \frac{A^2 (N+1) t_{cl}}{N} \Bigg[\frac{\sin{\omega t_{cl}/2}}{\omega t_{cl}}\Bigg]^2
\end{align}
where $\omega$ is the frequency, $t_{cl}$ is the clock period (minimum time between a change in levels) which is a multiple of the sampling time ($T_s$) and $A$ is the amplitude of the signal. Thus, for designing the PRBS signal it is necessary to estimate the amplitude and the frequency range.

\begin{align}
    \frac{2\pi}{T} \leq \omega \leq \frac{2.8}{t_{cl}}
\end{align}

\citeauthor{rivera1995systematic}, 1995, \citeauthor{lee2005integrated},2005 and \citeauthor{garciainput} provided practical guidelines for estimating the range of frequency needed for process closed-loop identification using time domain information. The primary frequency band of interest for excitation is determined by the dominant time constants of the system. 

\begin{align}
    \omega_{low} &= \frac{1}{S_f ~ t^{ol}}\\
    \omega_{high} &= \frac{4 S_f}{t^{cl}}\\
    \omega_{high} &\leq \omega_{N}
\end{align}

where $S_f$ is a safety factor used to augment the bandwidth of the excitation signal, $t^{ol}$ and $t^{cl}$ are the dominant time constants of the open loop and closed loop process respectively. Also, the upper value of the frequency must be lower than the Nyquist frequency $\omega_{N}$ to avoid aliasing. Although the magnitude of the signal has not been optimized in the current work, it could be further optimized by taking a profit function of the plant into consideration for minimal losses and using the validation data used for the FDD model.\\

\section{Results and discussion}
In this section, the industrial benchmark TEP is used to validate and demonstrate the effectiveness of the proposed method. We investigated the multi-class classification performance using a total of 20 fault modes which involve all of the compositions, manipulated and measurement variables in the TE process. For an individual class IDV(i), the performance was typically evaluated by a confusion matrix which consists of true positives (TP$_i$), false positives (FP$_i$), true negatives (TN$_i$) and false negatives (FN$_i$). The notation used in the confusion matrix is as follows:

\begin{table}[]
\begin{tabular}{@{}ccc@{}}
\toprule
 & \begin{tabular}[c]{@{}c@{}}Counts of \\ predicted label i\end{tabular} & \begin{tabular}[c]{@{}c@{}}Counts of  predicted \\ label other than i\end{tabular} \\ \midrule
Counts of real label $i$ & TP$_i$ & TN$_i$ \\
\begin{tabular}[c]{@{}c@{}}Counts of real label \\ other than $i$\end{tabular} & FP$_i$ & FN$_i$ \\ \bottomrule
\end{tabular}
\caption{Confusion Matrix for each fault (IDV(i))}
\end{table}

Two main important metrics for quantifying the performance of the proposed process monitoring methodology are as follows:
\begin{itemize}

\item Fault Detection Rate (FDR):
\begin{align}
\nonumber FDR &= \frac{
number(\text{fault data that have been detected as fault})}
{number(\text{fault data)}} \\
&= \frac{TP_i}{TP_i + FP_i}
\end{align}

FDR represents the probability that the abnormal conditions are correctly detected which is an important
criterion to compare between different methods in terms of their detection efficiency.
Evidently, a very high FDR is desirable.

\item  False Alarm Rate (FAR):

\begin{align}
\nonumber FAR &= \frac{number\text{(normal data that have been detected as fault)}}
{number(\text{normal data)}} \\ 
& = \frac{FP_i}{TP_i + TN_i}
\end{align}
where the class corresponding to normal operation is considered as the positive class. FAR represents the probability that the normal operation is wrongly identified as abnormal and thus a very low FAR is desired and necessary.
\end{itemize}

\begin{figure}
    \centering
    \includegraphics[scale = 0.65]{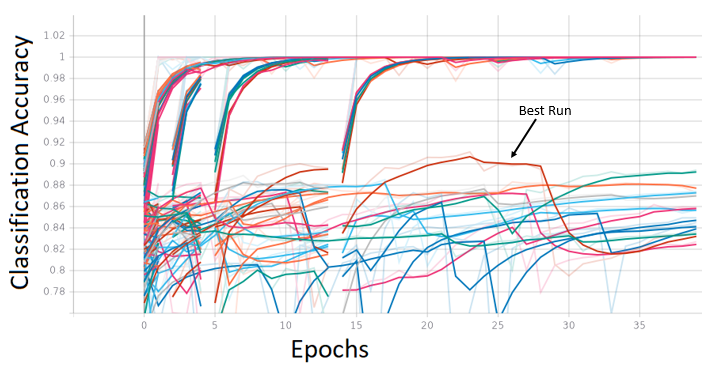}
    \caption{Training and Validation averaged classification accuracy for different values of hyper-parameters. Different colour represent different runs with a specific combination of hyper-parameters}
    \label{classification_training_validation}
\end{figure}

The fault detection results obtained with the RNN based model are compared with both linear multivariate statistical methods and deep learning methods reported in previous studies. For a fair comparison between the methods, for studies where only non-incipient faults were considered the results were compared to fault detection results obtained from the first level of the hierarchical structure model whereas for studies where all the faults were considered, the comparisons were done for results obtained from second level of the hierarchical structure model.  The fault detection rate (FDR) for all the faults is compared in Table 5 for the proposed method, PCA \cite{lv2016fault}, DPCA\cite{lv2016fault}, ICA\cite{hsu2010novel}, Convolutional NN (CNN) \cite{8926924}, Deep Stacked Network (DSN) \cite{chadha2017comparison}, Stacked Autoencoder (SAE) \cite{chadha2017comparison}, Generative Adversarial Network (GAN) \cite{spyridon2018generative} and One-Class SVM (OCSVM) \cite{spyridon2018generative}. The fault detection rates for all non-incipient faults and incipient faults are shown in Table \ref{Fault detection comparison} and \ref{Fault detection comparison1} respectively for different methodologies along with the results from the proposed method. It can been seen from Table \ref{Fault detection comparison} that the proposed method outperformed the linear multivariate methods and other DL based methods for most fault modes. For example, for PCA with 15 principal components, the average fault detection rates are 61.77\% and 74.72\% using $T^2$ and $Q$ statistic respectively. Since the principal components extracted using PCA captures static correlations between variables, DPCA is used to account for temporal correlations (both auto-correlations and cross-correlations) in the data. The effect of increasing the number of time samples in the Tennessee Eastman simulation is also investigated following the hypothesis that increasing the time horizon will enhance classification accuracy. In the case of DPCA, the number of lags used in the observation matrix is a key parameter. Since DPCA is only a data compression technique it must be combined with a classification model for the purpose of fault detection. Accordingly, the output features from the DPCA model are fed into an SVM model that is used for final classification. Different time horizons were tried for training the DPCA model. Based on validation results the best DPCA model  was obtained with 15 lags and thus this model is compared with an RNN also based on 15 lags. The average detection rate obtained was 72.35\%. ICA \cite{hsu2010novel} based monitoring scheme perform better than both PCA and DPCA based methods with an averaged accuracy of approximately 90\%. It should be noted that all these methods (PCA, DPCA and ICA) perform poorly for detecting incipient faults. 
In addition to the comparison to linear methods the proposed methodology was also compared with different DNN architectures such as CNN\cite{chadha2017comparison}, DSN \cite{chadha2017comparison}, SAE-NN (results reported in \citeauthor{chadha2017comparison},\citeyear{chadha2017comparison}) and GAN\cite{spyridon2018generative}, OCSVM (results reported in \citeauthor{spyridon2018generative},\citeyear{spyridon2018generative}) reported previously. It can be seen that the proposed method also outperforms these DNN based methodologies. The relative advantage of our method versus these other DNN architectures (Table 4) is mostly due to the inclusion of the incipient faults within the normal class. This reduces the confusion between the normal samples with other non-incipient faults. However, the additional advantage of the proposed method over the other DNN architectures is realized when the hierarchical structure is used in combination with the PRBS signals as further discussed below. It should be noted that all these comparisons were based on an identical data set.  
\\

\begin{figure}[]
    \centering
    \includegraphics[scale = 0.8]{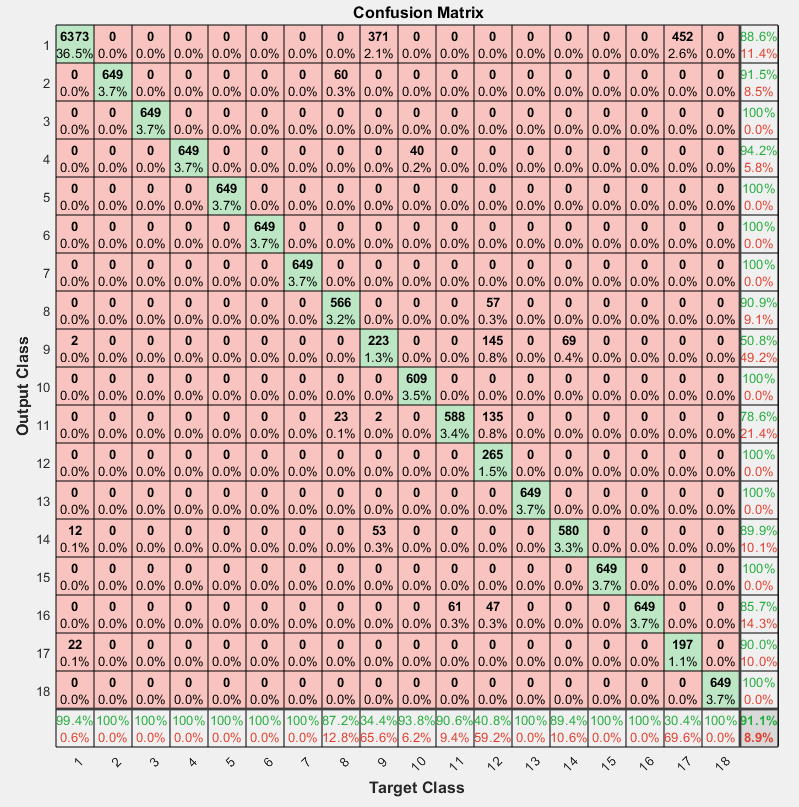}
    \caption{Confusion Matrix for the first level model of the hierarchical structure (i.e. classification of non-incipient faults and considering incipient faults as a normal class)}
    \label{level1_cm}
\end{figure}

 Subsequently, the faults were diagnosed using the proposed hierarchical structure where the first level model of the hierarchical structure classifies non-incipient faults and the second level model classifies incipient faults. For the first level model, there are 7382 training samples and 17,442 testing samples in total with a time horizon of 150 time-steps. The model consists of 182 encoder LSTM units, followed by 116 LSTM units for processing of the output of the encoding layer. Thereafter, the output of the second LSTM layer is passed through a dense layer for classification. Hyper-parameters such as number of layers, number of LSTM units in each layer, classification weights, learning rate, time-horizon etc.  are selected using validation data  that are part of the training dataset. The hyper-parameter search is implemented using keras-tuner. Firstly, a grid of hyper-parameters is defined, for example number of encoder layers = [1,2,3], number of LSTM units for each of these layers ranging from 2 to 200 with an interval of 2  = [10:2:200], learning rate =  [1e$^{-1}$,2e$^{-1}$,3e$^{-1}$, 1e$^{-2}$], value of weights in the objective function, etc. Keras-tuner trains the model using different combinations of these hyper-parameters values and the averaged validation accuracy is evaluated at every epoch. The models are trained with a few epochs in the start and the selected models with high validation accuracy are chosen to be trained for more epochs. A snapshot of the averaged training accuracy and validation accuracy is shown for different hyper-parameters in Figure \ref{classification_training_validation}. The marked region in Figure \ref{classification_training_validation} is the best run with highest validation accuracy and the combination of hyper-parameters for the run are used to evaluate test accuracy. A study was conducted to select the optimal time horizon for HDRNN based model. It can be seen from Figure \ref{comparison_time_horizon} that the classification averages can be enhanced by extending the length of the time horizon of past data fed to the RNN based model. 150 time steps were chosen as the optimal time-horizon. Confusion matrix for level 1 model is presented  in Figure \ref{level1_cm}.\\
 
 \begin{figure}[]
    \centering
    \includegraphics[scale = 0.5]{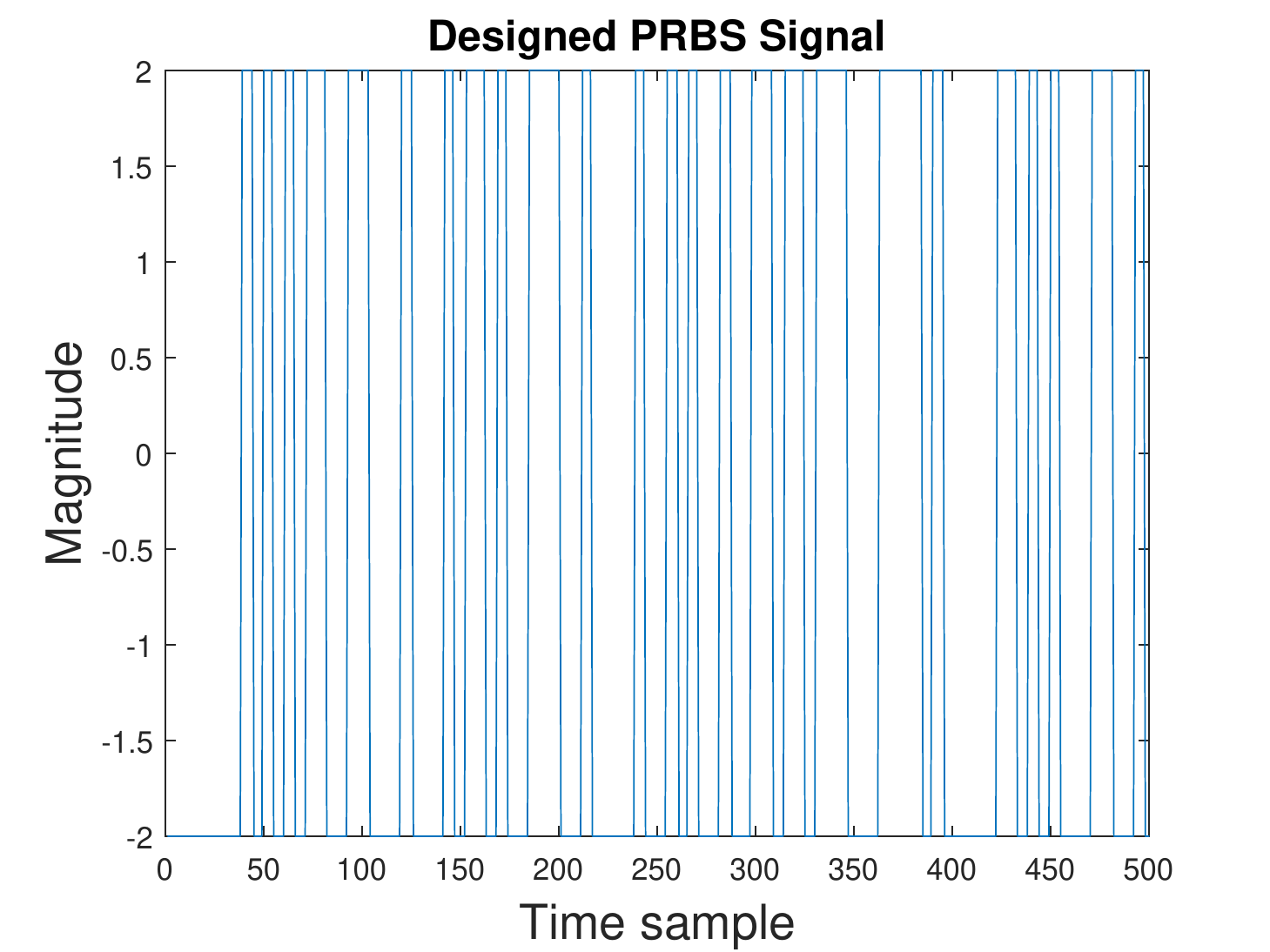}
    \includegraphics[scale = 0.5]{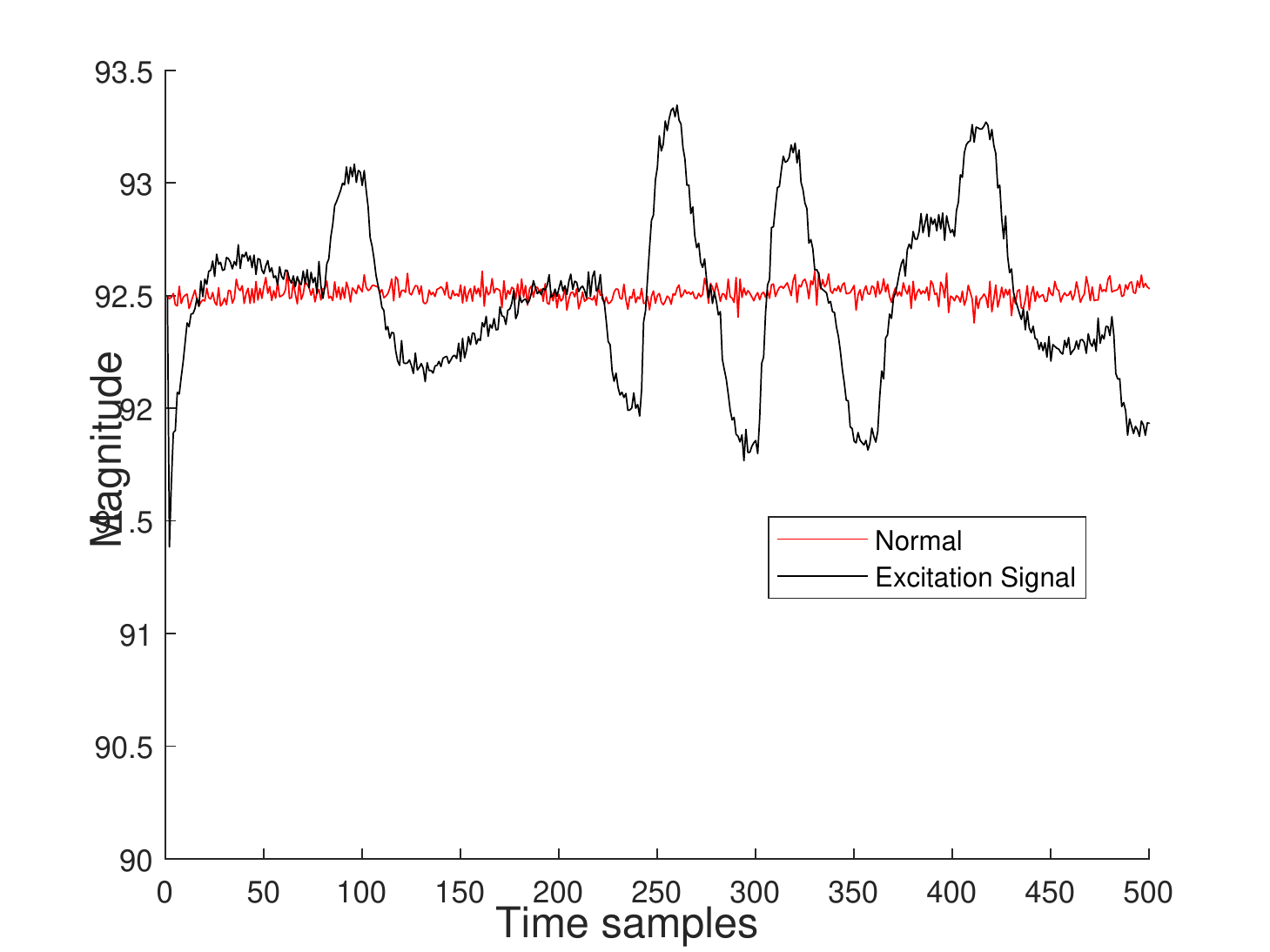}
    \caption{(a) Designed excitation signal for fault 15 (training dataset), (b) Output signal}
    \label{sep}
\end{figure}
 
 The next important design parameter for the second level hierarchical model is the location in the process at which the external excitation signal should be introduced to maximize information about the occurring incipient fault. In this work, this choice is based on the flow-sheet and by identifying which variables are mostly correlated to the incipient faults under consideration. Specifically, the excitation signals were added to process set-points in control loops that are most correlated to the incipient faults. When the selection of the variable to be excited by a PRBS is not obvious from the process flow-sheet, a more systematic approach is to use sensitivity analysis, e.g. sensitivity of changes in the variable connected to the fault to all process variables. Since it may be detrimental to perturb the set-point continuously by the PRBS signal the latter can be introduced intermittently into the process. In the current work an excitation signal of length 40 time-steps was intermittently introduced every 4 hours into the process by assuming that such event will not impact significantly the profitability of the process (for test data). Changes in the separator temperature set-point will force changes in the condenser temperature. Since the fault to be identified is stiction in the valve that affects the condenser temperature, the imposed PRBS in the separator set-point  indirectly helps in identifying fault 15.  A snapshot of the PRBS and the output signal is shown in Figure \ref{sep}. For fault 9 i.e. random variation in D feed temperature (refer Table \ref{process faults}) the PRBS excitation $(\omega \in [\omega_{cl},\omega_n]$ where $\omega_{cl} = 0.0087$ rad/s and  $\omega_n = 1.74$ rad/s) signal is introduced to the D feed ratio, in order to create a suitable excitation. After developing this PRBS signal, we added both signals to the process at different times during the simulation. For fault 15, the PRBS signal is designed with a frequency range of $\omega \in [\omega_{cl},\omega_n]$ where $\omega_{cl} = 0.005$ rad/s and  $\omega_n = 1.74$ rad/s\\
 
\begin{figure}[]
    \centering
    \includegraphics[scale = 0.6]{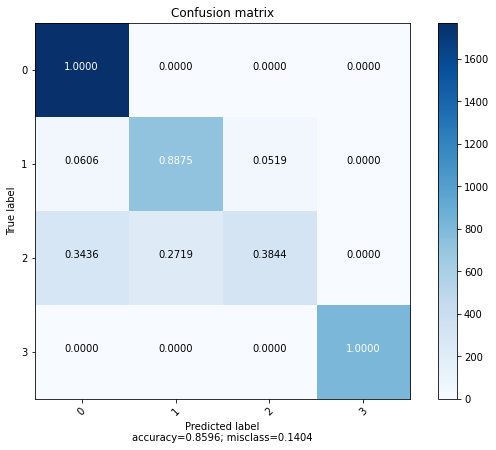}
    \includegraphics[scale = 0.6]{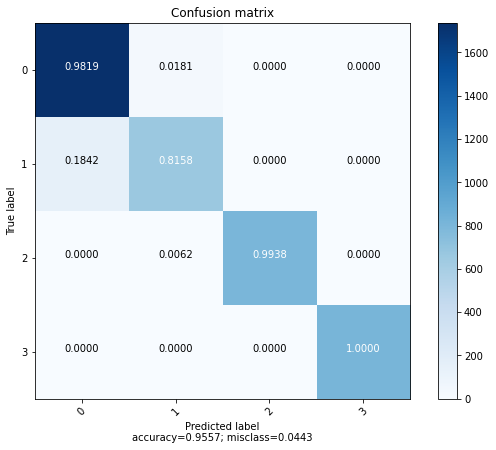}
    \caption{Confusion Matrix on test data for the second level model of the hierarchical structure: a) After adding designed PRBS signal w.r.t. fault 15 b) After adding designed PRBS signal w.r.t. fault 9 and fault 15 }
    \label{cm_level2}
\end{figure}

For the second level model, there are 1,796 training samples and 4,196 testing samples in total with a time horizon of 150 time-steps. The model consists of 284 encoder LSTM units in the first hidden layer, second layer consists of 100 LSTM units, followed by 278 LSTM units for processing of the output of the encoding layer. Thereafter, the output of the third LSTM layer is passed through a dense layer for classification. Hyper-parameters such as number of layers, number of LSTM units in each layer, classification weights, learning rate, time-horizon, weights in the loss function etc.  are selected using the validation data  which is part of the training dataset. The hyper-parameter search is implemented again using the keras-tuner. For the second level model, the samples corresponding to fault 0 (normal) and incipient faults are considered. Figure \ref{cm_level2} (a) shows the confusion matrix after introducing the PRBS signal that was designed for identifying fault 15 and Figure \ref{cm_level2} (b) shows the confusion matrix after introducing both PRBS signals that were designed for identifying fault 15 and fault 9. The total FAR calculated using Equation 17 was 2.41\%. \\

\begin{figure}[]
    \centering
    \includegraphics[scale = 0.5]{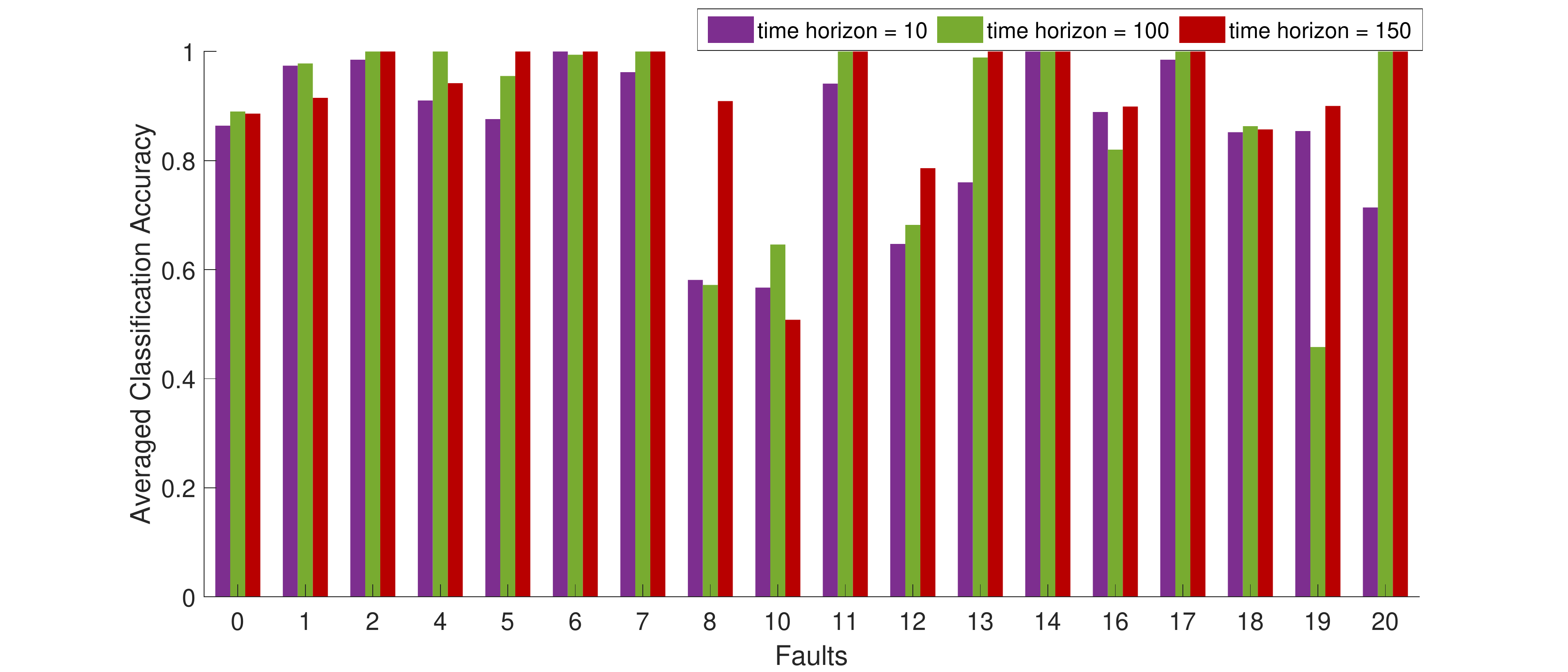}
    \caption{Selection of optimal time horizon for Hierarchical LSTM-SAE Level 1 model}
    \label{comparison_time_horizon}
\end{figure}

\begin{figure}
    \centering
    \includegraphics[scale = 0.5]{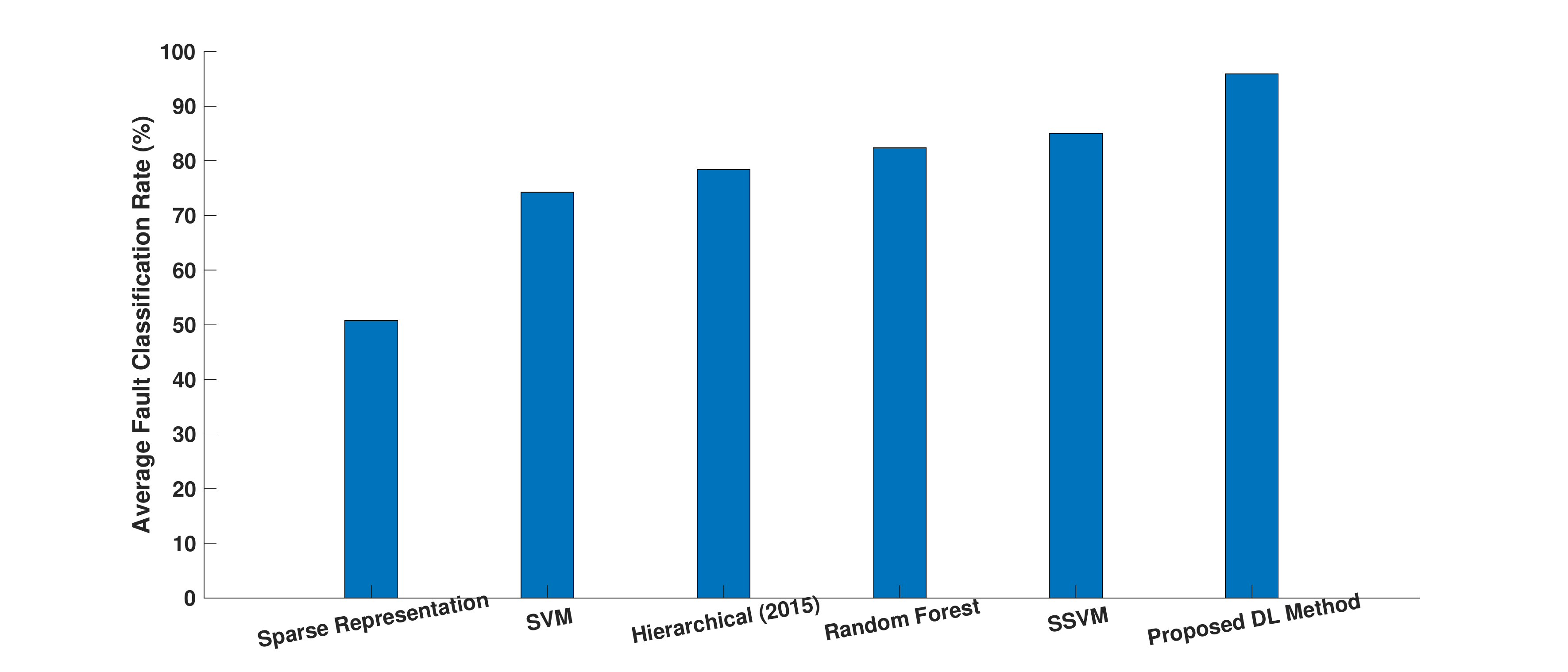}
    \caption{Comparison of averaged fault classification rates (non-incipient faults only)}
    \label{comparison_fault_classification}
\end{figure}

\begin{figure}
    \centering
    \includegraphics[scale = 0.5]{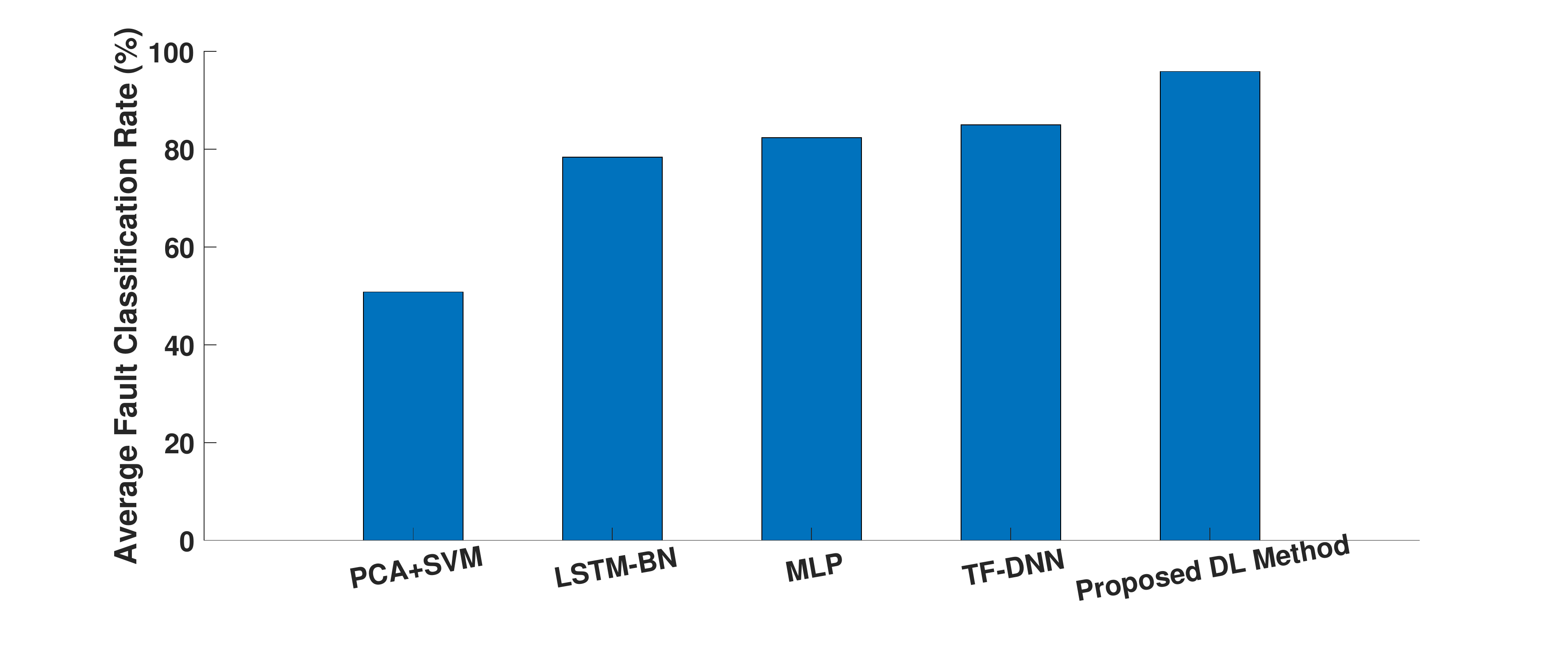}
    \caption{Comparison of averaged fault classification rates (all faults)}
    \label{comparison_fault_classification1}
\end{figure}

The averaged fault classification rate for all non-incipient faults and for all faults (including incipient faults) are shown in Figure \ref{comparison_fault_classification} and \ref{comparison_fault_classification1} respectively. Figure \ref{comparison_fault_classification} shows a bar-chart comparison of the proposed method with several non-linear methods such as Sparse representation\cite{wu2012fault}, SVM\cite{yin2014study}, Hierarchical model based method\cite{7424410}, Random Forest, Structural SVM. It can be seen that the Hierarchical Deep RNN based method outperforms other methods with a significant margin. It should be noted that comparisons made in Figure \ref{comparison_fault_classification} do not consider incipient faults. In Figure \ref{comparison_fault_classification1}, the averaged test accuracy of all faults (both incipient and non-incipient faults) are compared with other DL based methods\cite{luo2020deep}. It can be seen that the second level hierarchical model combined with the introduction of the designed PRBS signals significantly improves the classification of the incipient faults and thus the averaged test accuracy for fault diagnosis increases significantly.

\begin{table}[]
\caption{Comparison of Fault Detection Rate with different methods with non-incipient faults only}
\label{Fault detection comparison}
\resizebox{\textwidth}{!}{%
\begin{tabular}{@{}cccccccccccc@{}}
\toprule
\textbf{Fault} & \multicolumn{2}{c}{\textbf{\begin{tabular}[c]{@{}c@{}}PCA\\ (15 comp.)\end{tabular}}} & \textbf{\begin{tabular}[c]{@{}c@{}}DPCA\\ (22 comp.)\end{tabular}} & \multicolumn{2}{c}{\textbf{\begin{tabular}[c]{@{}c@{}}ICA\\ (9 comp.)\end{tabular}}} & \textbf{\begin{tabular}[c]{@{}c@{}}DL\\ (2017)\end{tabular}} &
\textbf{\begin{tabular}[c]{@{}c@{}}DL\\ (2017)\end{tabular}} &
\textbf{\begin{tabular}[c]{@{}c@{}}DL\\ (2018)\end{tabular}} &
\textbf{\begin{tabular}[c]{@{}c@{}}DL\\ (2018)\end{tabular}}&
\textbf{\begin{tabular}[c]{@{}c@{}}DL\\ (2019)\end{tabular}} & \textbf{\begin{tabular}[c]{@{}c@{}}Prop- \\osed DL \\ \end{tabular}} \\ \midrule
 & \textbf{$T^2$} & \textbf{SPE} & \textbf{$T^2$} & \textbf{$I^2$} & \textbf{AO} & {SAE-NN} & {DSN} & {GAN} & OCSVM & {CNN} &  HDRNN (LSTM-SAE)\\  \hline
1 & 99.2\% & 99.8\% & 99\% & 100\% & 100\% & 77.6\% & 90.8\% & 99.62\% & 99.5\%& 91.39\% & 100\% \\
2 & 98\% & 98.6\% & 98\% & 98\% & 98\% & 85\% & 89.6\% & 98.5\% & 98.5\% & 87.96\% & 100\% \\
4 & 4.4\% & 96.2\% & 26\% & 61\% & 84\% & 56.6\% & 47.6\% & 56.25\% & 50.37\% & 99.73\% & 100\% \\
5 & 22.5\% & 25.4\% & 36\% & 100\% & 100\% & 76\% & 31.6\% & 32.37\% & 30.5\% & 90.35\% & 100\% \\
6 & 98.9\% & 100\% & 100\% & 100\% & 100\% & 82.8\% & 91.6\% & 100\% & 100\%& 91.5\% & 100\% \\
7 & 91.5\% & 100\% & 100\% & 99\% & 100\% & 80.6\% & 91\% & 99.99\% & 99.62\%& 91.55\% & 100\% \\
8 & 96.6\% & 97.6\% & 98\% & 97\% & 97\% & 83\% & 90.2\% & 97.87\% & 97.37\% & 82.95\% & 100\% \\
10 & 33.4\% & 34.1\% & 55\% & 78\% & 82\% & 75.3\% & 63.2\% & 50.87\%& 53.25\% & 70.05\% & 42.84\% \\
11 & 20.6\% & 64.4\% & 48\% & 52\% & 70 & 75.9\% & 54.2\% & 58\% & 54.75\%& 60.16\% & 100\% \\
12 & 97.1\% & 97.5\% & 99\% & 99\% & 100\% & 83.3\% & 87.8\%& 98.75\% & 98.63\%& 85.56\% & 100\% \\
13 & 94\% & 95.5\% & 94\% & 94\% & 95\% & 83.3\% & 85.5\% & 95\% & 94.87\%& 46.92\% & 100\% \\
14 & 84.2\% & 100\% & 100\% & 100\% & 100\% & 77.8\% & 89\%& 100\% & 100 \%& 88.88\% & 100\% \\
16 & 16.6\% & 24.5\% & 49\% & 71\% & 78\% & 78.3\% & 74.8\%& 34.37\%& 36.37\% & 66.84\% & 100\% \\
17 & 74.1\% & 89.2\% & 82\% & 89\% & 94\% & 78\% & 83.3\% & 91.12\% & 87.25\%& 77.11\% & 100\% \\
18 & 88.7\% & 89.9\% & 90\% & 90\% & 90\% & 83.3\% & 82.4\% & 90.37\% & 90.12\% & 82.74\% & 100\% \\
19 & 0.4\% & 12.7\% & 3\% & 69\% & 80\% & 67.7\% & 52.4\%&  11.8\% & 3.75\%& 70.87\% & 40.4\% \\
20 & 29.9\% & 45\% & 53\% & 87\% & 91\% & 77.1\% & 44.1\%& 58.37\% & 52.75\% & 72.88\% & 100\% \\ \hline
Average & 61.77\% & 74.72\% & 72.35\% & 87.29\% & 91.70\% & 77.7\% & 76.84\% & 74.04\% &62.78\% & 85.47\% & 93.13\% \\ \bottomrule
\end{tabular}}
\end{table}

\begin{table}[]
\caption{Comparison of Fault Detection rate with different methods (with all faults)}
\label{Fault detection comparison1}
\resizebox{0.6\textwidth}{!}{%
\begin{tabular}{@{}cccccccccccc@{}}
\toprule
\textbf{Fault} &  \textbf{\begin{tabular}[c]{@{}c@{}}DL\\ (2017)\end{tabular}} &
\textbf{\begin{tabular}[c]{@{}c@{}}DL\\ (2017)\end{tabular}} &
\textbf{\begin{tabular}[c]{@{}c@{}}DL\\ (2018)\end{tabular}} &
\textbf{\begin{tabular}[c]{@{}c@{}}DL\\ (2018)\end{tabular}}&
\textbf{\begin{tabular}[c]{@{}c@{}}DL\\ (2019)\end{tabular}} & \textbf{\begin{tabular}[c]{@{}c@{}}Prop- \\osed DL \\ \end{tabular}} \\ \midrule
  & {SAE-NN} & {DSN} & {GAN} & OCSVM & {CNN} &  HDRNN (LSTM-SAE)\\  \hline
1  & 77.6\% & 90.8\% & 99.62\% & 99.5\%& 91.39\% & 100\% \\
2 & 85\% & 89.6\% & 98.5\% & 98.5\% & 87.96\% & 100\% \\
3 & 79.4\% & 14.4\% & 10.375\% & 7.62\% & 50.59\% & 81.58\% \\
4 & 56.6\% & 47.6\% & 56.25\% & 50.37\% & 99.73\% & 100\% \\
5 & 76\% & 31.6\% & 32.37\% & 30.5\% & 90.35\% & 100\% \\
6 & 82.8\% & 91.6\% & 100\% & 100\%& 91.5\% & 100\% \\
7  & 80.6\% & 91\% & 99.99\% & 99.62\%& 91.55\% & 100\% \\
8 & 83\% & 90.2\% & 97.87\% & 97.37\% & 82.95\% & 100\% \\
9 & 50.6\% & 16.3\% & 8.625\% & 7.125\%& 49.53\% & 99.38\% \\
10  & 75.3\% & 63.2\% & 50.87\%& 53.25\% & 70.05\% & 42.84\% \\
11 & 75.9\% & 54.2\% & 58\% & 54.75\%& 60.16\% & 100\% \\
12  & 83.3\% & 87.8\%& 98.75\% & 98.63\%& 85.56\% & 100\% \\
13 & 83.3\% & 85.5\% & 95\% & 94.87\%& 46.92\% & 100\% \\
14 & 77.8\% & 89\%& 100\% & 100 \%& 88.88\% & 100\% \\
15 & 55.5\% & 26.7\%& 12.5\% & 14\%& 43.54\% & 100\% \\
16 & 78.3\% & 74.8\%& 34.37\%& 36.37\% & 66.84\% & 100\% \\
17 & 78\% & 83.3\% & 91.12\% & 87.25\%& 77.11\% & 100\% \\
18 & 83.3\% & 82.4\% & 90.37\% & 90.12\% & 82.74\% & 100\% \\
19 & 67.7\% & 52.4\%&  11.8\% & 3.75\%& 70.87\% & 40.4\% \\
20 & 77.1\% & 44.1\%& 58.37\% & 52.75\% & 72.88\% & 100\% \\ \hline
Average & 75.355\% &65.32\% & 64.51\% &62.78\% & 79.84\% & 93.23\% \\ \bottomrule
\end{tabular}}
\end{table}

\section{Conclusions}
This work studied the application of a deep learning model within a hierarchical structure as a way to increase the detection and classification of faults in the Tennessee Eastman Process (TEP). The TEP simulation contains 20 different faults that were used during this study to make the classification problem. As previously reported by other researchers a subset of these faults, referred in these study as incipient, are particularly difficult to diagnose due to low signal to noise ratio and similarities in the resulting dynamic responses corresponding to different faults. \\

A comparison between deep learning techniques to a multivariate linear technique for fault detection such as PCA, DPCA, ICA and other deep learning methods is also presented. It is observed that RNN-Hierarchical based model is superior than traditional linear and other deep learning based methods for fault classification due to their ability to capture nonlinear dynamic behaviour. It was also shown that the classification averages can be enhanced by extending the length of the time horizon of past data fed to the RNN based model. However, most of these improvements in classification occurred for the non-incipient faults. Therefore, an active fault detection approach was pursued where a hierarchical model structure combined with external PRBS signals was proposed that proved to be particularly effective for classifying incipient faults.

\section{Acknowledgement}
This work is the result of the research project supported by MITACS grant IT10393 through MITACS-Accelerate Program.








\bibliographystyle{cas-model2-names}

\nocite{*}
\bibliography{cas-refs.bib}
\end{document}